# Analysis and Representation of Igbo Text Document for a Text-Based System

Ifeanyi-Reuben Nkechi J.[1], Ugwu Chidiebere[2], Adegbola Tunde[3]
[1]Department of Computer Science, Rhema University, Nigeria
[2]Department of Computer Science, University of Port Harcourt, Nigeria
[3]African Language Technology Initiative (ALT-i) Ibadan, Nigeria
Email: ifyreubennkechi@yahoo.com, chidiebere.ugwu@uniport.edu.ng, taintransit@yahoo.com

**Abstract** — The advancement in Information Technology (IT) has assisted in inculcating the three Nigeria major languages in text-based application such as text mining, information retrieval and natural language processing. The interest of this paper is the Igbo language, which uses compounding as a common type of word formation and as well has many vocabularies of compound words. The issues of collocation, word ordering and compounding play high role in Igbo language. The ambiguity in dealing with these compound words has made the representation of Igbo language text document very difficult because this cannot be addressed using the most common and standard approach of the Bag-Of-Words (BOW) model of text representation, which ignores the word order and relation. However, this cause for a concern and the need to develop an improved model to capture this situation. This paper presents the analysis of Igbo language text document, considering its compounding nature and describes its representation with the Word-based N-gram model to properly prepare it for any text-based application. The result shows that Bigram and Trigram n-gram text representation models provide more semantic information as well addresses the issues of compounding, word ordering and collocations which are the major language peculiarities in Igbo. They are likely to give better performance when used in any Igbo text-based system.

## I. INTRODUCTION

Igbo language is one of the three main and official languages spoken in Nigeria [1] and mainly spoken in the south eastern part of the country. The Information Technology (IT) has evolved to the extent of using this language for its operation and creation of data. One can operate (Windows 7 and above) operating system, Microsoft office package and create documents using this language. It is envisaged that in future, as more people are getting used to the new trend, there will likely be large textual data generated with this language together with the growth of Igbo online information, which needs to be managed efficiently. These foreseen challenges have motivated the researcher to introduce an efficient computational approach to represent the language text document for any text-based system like text mining system, information retrieval system, and natural language processing system, owing to the fact that the text representation is one of the key essentials in text-based system. Text representation is a major basic and vital task in any text-based intelligent system. It involves looking for suitable terms to transfer text documents into numerical vectors [2].

Igbo language is an agglutinative language, the language in which words are built up stringing different words. The individual meaning of words in a phrase or compound word does not entail the context it is been used for. The text document representation is one of the issues resulting from natural language peculiarities that need to be resolved for the success of any research in the text related fields. According to Soumya G.K. and Shibily J.[3], Bag-Of-Words (BOW) model is a traditional and conventional approach for representing text documents in any text-based task. The BOW model treats words independently and does not consider the word ordering, compound words, and collocations for semantic enhancement. When this model is used in an Igbo text document, it will not be dully represented because of compound nature of the language. In Igbo language, compounding is a common type of word formation and many compound words exist. Compound words play high roles in the language. They can be referred as Igbo phrases that make sense only if considered as a whole. Examples are "ụlọ akwụkwọ - school"; "onye nkuzi – teacher"; "kọmputa nkunaka – laptop". Majority of Igbo terms, key words or features are in phrasal structure. The semantic of a whole is not equal to the semantic of a part [4]. Some online translators in Igbo language is entangled with this problem of improper text representation. For example if you want to translate "school" in Igbo using most of the systems, it will either display "ụlọ" or "akwụkwọ" which is wrong. This is because BOW model is used in text representation and did not capture the compound word "ụlọ akwụkwọ" as a whole it should be in Igbo for "school".

This paper presents an analysis of Igbo compound word and proposes an efficient computational approach for representing the text for any text-based work using word-based N-gram model. This work is a part of ongoing research to create a text classification system for a text documents in an Igbo language. The model chosen by the researchers will help to discover unidentified facts and concealed knowledge that may exist in the lexical, semantic or relations [4] in Igbo text corpus. In Igbo, the semantic of individual words in a compound and combinatory semantics of the larger units are not the same at all [5].

## II. IGBO LANGUAGE

A language is a method of communication between individuals who share common code, in form of symbols [5]. The Igbo language is one of the three major languages (Hausa, Yoruba and Igbo) in Nigeria. It is largely spoken by the people in the eastern part of Nigeria. Igbo language has many dialects. The standard Igbo is used formally and is adopted for this research. The current Igbo orthography (Onwu Orthography, 1961) is based on the Standard Igbo. Orthography is a way of writing sentence or constructing grammar in a language. Standard Igbo has thirty-six (36) alphabets (a, b, ch, d, e, f, g, gb, gh, gw, h, i, ị, j, k, kw, kp, l, m, n, nw, ny, ṅ, o, ọ, p, r, s, sh, t, u, ụ, v, w, y, z), consisting of eight (8) vowels and twenty-eight (28) consonants. The 28 consonant characters are "b, ch, d, f, g, gb, gh, gw, h, j, k, kw, kp, l, m, n, nw, ny, ṅ, p, r, s, sh, t, v, w, y, z" and 8 vowels characters are "a, e, i, ị, o, ọ, u, ụ". There are nine consonants characters that are digraphs: "ch, gb, gh, gw, kp, kw, nw, ny, sh" [6]. It uses a Roman Script and it is a tonal language with two distinct tones, high and low. Igbo is an agglutinative language, in which words are built by stringing different morphemes or words together [5]. Igbo language has a large number of compound words. A compound word is a word that has more than one root, and can made from combination of either nouns, pronouns or adjectives.

### III. ANALYSIS OF IGBO COMPOUND WORDS

A compound word is a combination of two or more words that work as a single new unit with a new meaning. It consists of two or more words which are capable of independent existence [5]. A compound word contains more than one root-word. We used the concept of Onukawa M.C. [7] to study Igbo compound words. They are analyzed and categorized as follows:

*Nominal (NN) Compound Word:* A nominal compound word is formed by the combination of two or more nouns. The nominal compound words are written separately not minding the semantic status of the nouns in Igbo. Example of Igbo nominal compound words are: nwa akwụkwọ - student; onye nkuzi – teacher; ama egwuregwu – stadium; ụlọ ọgwụ - hospital; ụlọ akwụkwọ - school.

*Agentive Compound Words*: In agentive compound word, one or more nouns express the meaning of the agent, doer of the action. The Igbo agentive compound words are written separately irrespective of the translations in English. They can also be referred to as VN (Verb Noun) compound words. Example: oje ozi – messenger; oti igba - drummer.

*Igbo Duplicated Compound Word*: Igbo duplicated compound words are formed by the repetition of the exact word two or more times to show a variety of meaning [7]. For example: ọsọ ọsọ - quickly; mmiri mmiri – watery; ọbara ọbara – reddish.

*Igbo Coordinate Compound Words*: This compound word is formed by the combination of two or words joined by the Igbo conjunction "na" meaning "and" in English. All the Igbo compound words of this category is written separately. Example: Ezi na ụlọ - family; okwu na ụka – quarrel.

*Igbo Proper Compound Words*: This category of Igbo compound words includes personal names, place names, and club names. All words in this category are wriiten together not minding how long they may be. Example: Uchechukwu; Ngozichukwuka; Ifeanyichukwu.

*Igbo Derived Compound Words*: The derived Igbo compound words are words derived from verbs or phrases. The roots of the derived Igbo compound words are written together. Example: Dinweụlọ - landlord. Igbo, being an agglutinative language, has a huge number of compounds words and can be referred to as a language of compound words. The summary of the analysed and categorised Igbo compound word is shown in TABLEI.

### IV. TEXT REPRESENTATION

Text representation is the selection of appropriate features to represent document [8]. The approach in which text is represented has a big effect in the performance of any text-based applications [9]. It is strongly influenced by the language of the text. Vector Space Model (VSM) is the most commonly used text representation model [10]. The model used BOW model in representing its terms or features. It involves computing word frequency in the text document and creates a vector in the form of features, which is equivalent to the words in the text. This text representation model cannot represent Igbo text effectively; the issues of collocations, compounding and word ordering that are well acquainted with Igbo languages will not be dully captured. This paper introduces using word-based N-gram model to represent Igbo language text document for use by any text-based system to address issues the language agglutinative nature which is not captured using the common text representation model, BOW. This involves representing the text in sequence of words, thereby strengthening the semantic relationship among words in the text.

Table 1 : Igbo Compound Words

| Igbo Compound Words | Meaning | Roots and meaning | Compound Word Category |
|---|---|---|---|
| Onye nkuzi | Teacher | Onye – Person Nkuzi – Teach | Nominal |
| Ezi na ụlọ | Family | Ezi – surrounding Na – and ụlọ - family | Coordinate |
| Ojiiegoachọego | businessman | Ojiiego – use money achọego – find money | Derived |
| ụgbọ ala | Car, motor | ụgbọ - vessel ala - land (road) | Nominal |
| Egbe igwe | Thunder | Egbe – gun Igwe – sky | Nominal |
| Iri abụọ | Twenty | Iri – ten Abụọ - two | Nominal |
| Ode akwụkwọ | Secretary | Ode – Write Akwụkwọ - book | Agentive |
| Eberechukwu | God's mercy | Ebere – mercy Chukwu - God | Proper |
| Mmiri mmiri | Watery | Mmiri -water | Duplicate |
| ọcha ọcha | Whitish | ọcha – white | Duplicate |
| Onye nchekwa | Administrator | Onye – person Nchekwa – protect | Nominal |
| Kọmputa Nkunaka | Laptop | Kọmputa – Computer Nkunaka – Handcarry | Nominal |
| Ọkpụ ụzụ | Blacksmith | ọkpụ - mold ụzụ - clay | Agentive |
| Nche anwụ | Umbrella | Nche – protect | Agentive |

| | | Anwụ - sun | |
|---|---|---|---|
| Onyonyo kọmputa | Monitor | Onyonyo – screen Kọmputacomputer | Nominal |
| Okwu ntughe | Password | Okwu – speech Ntughe - opening | Nominal |

### V. RELATED WORKS

Wen Zhang et al [2] studied and compared the performance of adopting TF*IDF, LSI (Latent Semantic Indexing) together with multiple words for text representation. They used Chinese and English corpora to assess the three techniques in information retrieval and text categorization. Their result showed that LSI produced greatest performance in retrieving English documents and also produced best performance for Chinese text categorization. Chil-Fong Tsai [11] improved and applied BOW for image annotation. An image annotation is used to allocate keywords to images automatically and the images are represented using characteristics such as color, texture and shape. This is applied in Content-Based Image Retrieval System (CBIRS) and the retrieval of the image is based on indexed image features.

Don Shen et al [8] proposed an n-multigram language model approach to represent document. This approach can automatically determine the unknown semantic order of words in a document under a given category. The experiment shows that n-multigram model can achieve a similar result with n-gram model. Wanjiku Ngàngà emphasized that the availability of electronic resources for textual documents of various languages online contributes immensely to the language technology research and development. His paper described the methodology for creating machine-readable Igbo dialectal dictionary from an audio corpus rather than text corpora [12]. In 2004, Harmain et al [13] presented the architecture of Arabic text mining system as well as some issues involved in its mining task. This started with a preprocessing task, which converted the Arabic HTML documents to XML documents. The preprocessed text is then analyzed based on linguistic factors from the word level to the text level. The output of the analysis represented as a semantic network of the entities of the text and the relationships among them is established. This semantic network is then made available for some particular mining tasks. These related research works are mainly on English language; European languages (like French, German and Spanish); Asian languages (like Chinese and Japanese) and Arabic language. Slight or no work on this research area has been done on Igbo, a Nigerian language. Thus, this work will study, analyse the Igbo text document considering the compounding nature of its word formation and introduces a better text representation approach that can be adopted to improve the performance of its text-based system.

### VI. METHODOLOGY

The bulk of concerns for any text-based system are attributed to text representation considering the peculiarities of the natural language involved. In this section, we propose an efficient and effective model to represent Igbo text to be adopted by any text-based system. This is a process of transforming unstructured Igbo textual document into a form proper for automatic processing. This is a vital step in text processing because it affects the general performance of the system. The proposed approach for the Igbo text representation process is shown in Fig. 1.

*A. Igbo Text*
The Unicode model was used for extracting and processing Igbo texts from file because it is one of the languages that employ non-ASCII character sets like English. Unicode supports many character sets. Each character of the characters in the set is given a number called a code point. This enabled us manipulate the Igbo text loaded from a file just like any other normal text. When Unicode characters are stored in a file, they are encoded as a stream of bytes. They only support a small subset of Unicode. Processing Igbo text needs UTF-8 encoding. UTF-8 makes use of multiple bytes and represents complete collection of Unicode characters. This is achieved with the mechanisms of decoding and encoding. Decoding translates text in files in a particular encoding like the Igbo text written with Igbo character sets into Unicode while encoding write Unicode to a file and convert it into an appropriate encoding [12]. We achieved this with the help of Python program and other Natural Language Processing tools. The mechanism is illustrated in Fig. 2.

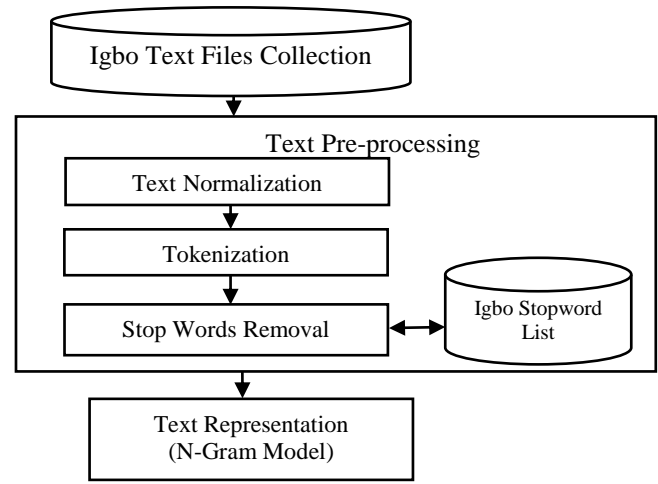

Fig. 1.  Igbo Text Representation Process

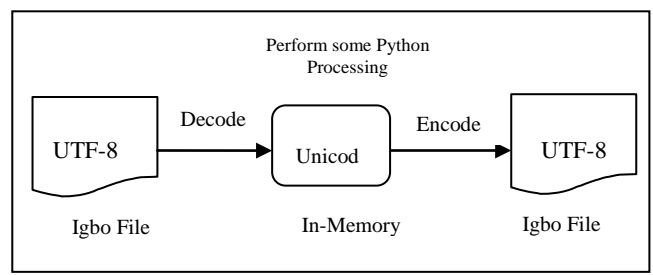

Fig. 2.  Igbo Text Unicode Decoding and Encoding

*B. Text Preprocessing*
Text preprocessing involves tasks that are performed on text to convert the original natural language text to a structure ready for processing. It performs very important functions in different text-based system. The tasks are Igbo text normalization, Igbo text tokenization and Igbo text Stop words Removal.
*B1. Igbo Text Normalization:* In Normalization process, we transformed the Igbo textual document to a format to make its contents consistent, convenient and full words for an efficient processing. We transformed all text cases to lower case and also removed diacritics and noisy data. The noisy data is assumed to be data that are not in Igbo dataset and can be from:

i. Numbers: Numbers can be cardinal numbers (single digits: 0-9 or a sequence of digits not starting with 0); signed numbers (contains a sign character (+ or -) following cardinal numbers immediately).
ii. Currency: This is mostly symbols used for currency e.g. £ (pound sign), € (Euro sign), ₦ (Naira sign), $ (dollar sign).
iii. Date and Time
iv. Other symbols like punctuation marks (:, ;, ?, !, ' ), and special characters like <, >, /, @, ", !, *, =, ^, %, and others.

A list is created for these data and the normalization task process is shown in Fig.3 and is performed before text tokenization.

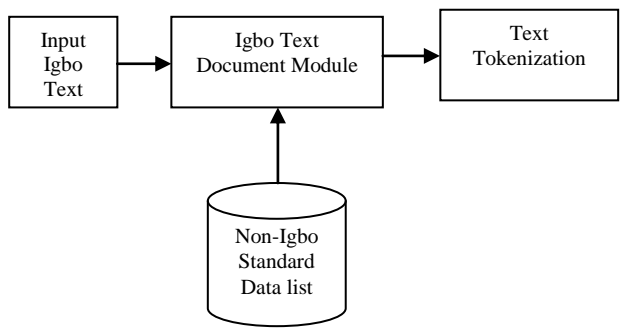

Fig. 3. Text Normalization Process

An algorithm designed for the text normalization task is given in algorithm 1.

Algorithm 1**:** Designed Igbo Text Normalization Algorithm.
Input: Igbo Text Document, Non-Igbo Standard Data/Character list
Output: Normalized Igbo Text

Procedure:
1. Transform all text cases to lower case.
2. Remove diacritics (characters like ū, ù, and ú contains diacritics called tone marks).
3. Remove non-Igbo standard data / character:
4. 
For every word in the Text Document,
  i. IF the word is a digits (0, 1, 2, 3, 4, 5, 6, 7, 8, 9) or contains digits THEN the word is not useful, remove it.
  ii. If the word is a special character (:, ;, ?, !, ', (, ), {, }, +, &, [, ], <, >, /, @, ", !, *, =, ^, %, and others ) or contains special character, the word is non-Igbo, filter it out.
  iii. If the word is combined with hyphen like "nje-ozi", "na-aga", then remove hyphen and separate the words. For example, the following word "nje-ozi" will be nje ozi, two different words.
  iv. If the word contains apostrophe like n'elu, n'ụlọ akwụkwọ then remove the apostrophe and separate

the words. For example "n'ụlọ akwụkwọ, after normalization will be three words n, ụlọ and akwụkwọ.

*B2    Text Tokenization:* Tokenization is the task of analyzing or separating text into a sequence of discrete tokens (words). The tokenization procedure for the work is shown in algorithm 2.
Algorithm 2:    Algorithm to tokenize the Igbo text
1. Create a TokenList.
2. Add to the TokenList any token found.
3. Separate characters or words, if the string matches any of the following: "ga-", "aga-", "n'", "na-", "ana-", "ọga-", "ịga-", "ọna-", "ịna-". For instance, the following strings: n'aka, na–ese, ịna-eche, ana-eme in a document will be separated into n', aka, na-, ese, ina-, eche, ana-, eme tokens.
4. Remove diacritics. This involves any non-zero length sequence of a–z, with grave accent (`), or acute accent (´), for example, these words ìhè and ájá appearing in a given corpus will be taken as ihe and aja tokens, removing their diacritics.
5. Any string separated with a whitespace is a token.
6. Any single string enclosed in double or single quote" should be treated as a token.
7. Any single string that ends with comma (,) or colon (:) or semi-colon (;) or exclamation mark (!) or question mark (?) or dot (.), should be treated as a token.

*B3    Igbo Stop-words Removal:* Stop-words are language-specific functional words; the most frequently used words in a language that usually carry no information [13].There are no specific amount of stop-words which all Natural language processing (NLP) tools should have.
Most of the language stop-words are generally pronouns, prepositions, and conjunctions. This task removes the stopwords in Igbo text. Some of Igbo stopwords is shown in Fig. 4.

> ndị, nke, a, i, ị, o, ọ, na, bụ, m, mụ, ma, ha, ụnụ, ya, anyị, gị, niine, nile, ngị, ahụ, dum, niile, ga, ka, mana, maka, makana, tupu, e, kwa, nta, naanị, ugbua, olee, otu, abụọ, atọ, anọ, ise, isii, asaa, asatọ, iteghete, iri, anyị, ndị, a, n', g', ụfọdu, nari, puku

Fig. 4. Sample of Igbo Stop-words list

In the proposed system, a stop-word list is created and saved in a file named "stop-words" and is loaded to the system whenever the task is asked to perform. The removal of the stop words in the proposed system is done following the designed algorithm 3.
Algorithm 3:    Algorithm to Remove Igbo Stop Words
1. Read the stop-word file
2. Convert all loaded stop words to lower case.
3. Read each word in the created Token List
4. For each word w ϵ Token List of the document
  i. Check if w(Token List) is in Language stop word list
  ii. Yes, remove w(Token List) from the Token List
  iii. Decrement tokens count
  iv. Move to the next w(Token List)

v. No, move to the next w(Token List)
vi. Exit Iteration Loop
vii. Do the next task in the pre-processing process
5. Remove any word with less than three (3) character length.

*C. N-Gram Based Text Representation*

In the N-Gram representation of text in this research, the "N" will span across 1 to 3, that means we will represent the text using Unigram when N=1; Bigram when N=2 and Trigram when N=3. The document is represented by word-based n-gram model. Given document j characterised by $d_j$, $f_{ij}$ is frequency of n-gram words $nw_i$ in the document $j$. The sample of Igbo text document that is used for the illustration of the proposed approach for the representation of Igbo text for any text applications is given in Doc1.

Igbo TextDoc1: Kpaacharụ anya makana prọjektọ nkụziihe a achọghị okwu ntụghe, ndị ichọghị ka ha hụ ga ahụ ihe-ngosi gi. Ọbụrụ na ichọrọ iji prọjektọ nkụziihe a were rụọ ọrụ, pịkịnye "Jikọọ". A na-akwụnye prọjektọ nkụziihe na kọmputa nkunaka iji mee ihe onyonyo. kọmputa nkunaka banyere na prọjektọ nkụziihe ọcha.

C1.Unigram Text Representation
A unigram model represents words or terms independently. It represents document in single words. Unigram probability estimates possibility of a word existing in text / corpora based on its general frequency of occurrence estimated from a corpus. Unigram model is based on BOW model and adopts likely method with BOW model in text representation and give the same result [3] and is shown in TABLEII.

Table 1: Unigram Representation of Igbo Text Doc1

| Unigram | Freq | Unigram | Freq |
|---|---|---|---|
| achọghị | 1 | nkụziihe | 4 |
| anya | 1 | ntụghe | 1 |
| banyere | 1 | okwu | 1 |
| gi | 1 | onyonyo | 1 |
| hụ | 1 | prọjektọ | 4 |
| ihe | 1 | pịkịnye | 1 |
| ihe-ngosi | 1 | rụọ | 1 |
| iji | 2 | were | 1 |
| jikọọ | 1 | ichọghị | 1 |
| kọmputa | 2 | ichọrọ | 1 |
| mee | 1 | ọbụrụ | 1 |
| na-akwụnye | 1 | ọcha | 1 |
| nkunaka | 2 | ọrụ | 1 |
| kpaacharụ | 1 | | |

Definition 1: The probability of representing text in unigram language model is approximated as follows:

$P(w_1, w_2, ..., w_T) \approx P(w_1) P(w_2) ... P(w_T)$ ..........................(1)

This implies that $P(w_1, w_2, ..., w_T) \approx \prod P(w_i)$ for i = 1,2,...T
For example: P(w=ụlọ akwụkwọ) = P(ụlọ) * P(akwụkwọ)

Figure 6 shows the unigram representation of Igbo text.

C2.Bigram Text Representation
Bigram probability estimates the possibility a word existing in the context of a previous word that means the probability of occurrence of a word lies on the probability of the previous word.

Definition 2: For a given word order W={w1,w2, ...wn}, the probability of a word order in bigram language model (n=2) is given as $P(w_1,w_2,...,w_n) = P(w_1 | w_2,w_3,...,w_n)P(w_2 | w_3,...,w_n) ... P(w_{n-1} | w_n)P(w_n)$ ........ (2)

C3.Trigram Text Representation
Trigram language model (n=3) uses the most recent two words of the document to condition the probability of the next word.

Definition 3: The probability of a term (word) sequence in trigram is
$P(w_3 | w_1, w_2)$ = count $(w_1, w_2, w_3)$ / count $(w_1, w_2)$ ..................................... (3)
Where count $(w_1, w_2, w_3)$ is the frequency of the order (sequence) of words $\{w_1, w_2, w_3\}$ and count $(w_1, w_2)$ is the frequency of the word order $\{w_1, w_2\}$ in the document.

Table 2: Bigram Representation of Igbo TextDoc1

| Bigram | Freq | Bigram | Freq |
|---|---|---|---|
| prọjektọ nkụziihe | 4 | kpaacharụ anya | 1 |
| kọmputa nkunaka | 2 | anya prọjektọ | 1 |
| achọghị okwu | 1 | ọrụ pịkịnye | 1 |
| na-akwụnye prọjektọ | 1 | nkunaka iji | 1 |
| ọbụrụ ichọrọ | 1 | ichọghị hụ | 1 |
| iji mee | 1 | rụọ ọrụ | 1 |
| were rụọ | 1 | nkụziihe kọmputa | 1 |
| ihe onyonyo | 1 | onyonyo kọmputa | 1 |
| nkunaka banyere | 1 | jikọọ na-akwụnye | 1 |
| okwu ntụghe | 1 | mee ihe | 1 |
| nkụziihe were | 1 | hụ ihe-ngosi | 1 |
| pịkịnye jikọọ | 1 | nkụziihe achọghị | 1 |
| banyere prọjektọ | 1 | ntụghe ichọghị | 1 |
| nkụziihe ọcha | 1 | gi ọbụrụ | 1 |
| iji prọjektọ | 1 | ihe-ngosi gi | 1 |

Table 3: Trigram Representation of Igbo TextDoc1

| Trigram | Freq | Trigram | Freq |
|---|---|---|---|
| ntụghe ichọghị hụ | 1 | nkunaka banyere prọjektọ | 1 |
| prọjektọ nkụziihe were | 1 | ọrụ pịkịnye jikọọ | 1 |
| prọjektọ nkụziihe kọmputa | 1 | ichọrọ iji prọjektọ | 1 |
| gi ọbụrụ ichọrọ | 1 | kọmputa nkunaka banyere | 1 |
| ihe onyonyo kọmputa | 1 | nkunaka iji mee | 1 |
| prọjektọ nkụziihe ọcha | 1 | pịkịnye jikọọ | 1 |
| | | na-akwụnye | |
| okwu ntụghe ịchọghị | 1 | rụọ ọrụ pikinye | 1 |
| nkụziihe kọmputa nkunaka | 1 | onyonyo kọmputa nkunaka | 1 |
| kọmputa nkunaka iji | 1 | hụ ihe-ngosi gi | 1 |
| jikọọ na-akwụnye projektọ | 1 | iji mee ihe | 1 |
| achọghị okwu ntụghe | 1 | nkụziihe were rụọ | 1 |
| ihe-ngosi gi ọbụrụ | 1 | banyere projektọ nkụziihe | 1 |
| were rụọ ọrụ | 1 | nkụziihe achọghị okwu | 1 |
| ịchọghị hụ ihengosi | 1 | kpaacharụ anya projektọ | 1 |
| anya projektọ nkụziihe | 1 | projektọ nkụziihe achọghị | 1 |
| ọbụrụ ịchọrọ iji | 1 | na-akwụnye projektọ nkụziihe | 1 |
| mee ihe onyonyo | 1 | iji projektọ nkụziihe | 1 |

## VII. RESULT

We have studied, analysed and represented Igbo textual document with word-based n-gram model and the results are shown in TABLES II, III and IV which infers for unigram model, bigram model and trigram model respectively that key features in the sampled Igbo text are not captured with unigram text representation model. TABLE II shows the result of unigram features generated from the Igbo text document used for the analysis, which is equivalent to the features extracted with BOW model. Only two key features "kọmputa – Computer" and "projektọ - Projector" but the context in which the word is used is not dully represented. TABLE III shows the results of features obtained with the bigram model. The model was able to extract bigrams that accurately described some concepts and compound words in Igbo language such as kọmputa nkunaka - laptop. TABLE IV displays result obtained with the trigram model of text representation in Igbo text document used for the analysis. TABLEV shows the summary of key features obtained from the results shown in TABLES II, III and IV.

Table 4: Igbo TextDoc1 N-Gram key Features

| Igbo TextDoc1 | | |
|---|---|---|
| Unigram Key Features | Bigram Key Features | Trigram Key Features |
| Kọmputa - Computer | Projektọ Nkụziihe | Projektọ nkuziihe kọmputa |
| Projektọ - Projector | Kọmputa nkunaka | Ihe onyonyo kọmputa |
| | Okwu ntụghe | Onyonyo kọmputa nkunaka |
| | Ihe onyonyo | |
| | Onyonyo kọmputa | |
| 2 | 5 | 3 |

From our observation and analysis of the results obtained, semantically, unigram model / BOW is not an ideal model for representing Igbo textual document in any text-based application. Bigram and Trigram text representation models provide more semantic information as well addresses the issues of compounding, word ordering and collocations which are the major language peculiarities in Igbo; that play high role in the language. They highly give an idea of context a word is used, thus increasing the quality of features. They are likely to give better performance when used in any Igbo text-based application. For example, from the sample Igbo TextDoc1, the compound words "kọmputa nkunaka – laptop" and "okwu ntụghe – password" are dully represented with bigram model. And the compound word "Ihe Onyonyo Kọmputa – Computer Screen" is also dully and semantically represented with trigram model.

## VIII. CONCLUSION

An improved computational approach for representing Igbo text document using word-based n-gram model considering the agglutinative nature of Igbo language has been developed. The result has shown that Bigram and Trigram are novel approaches for representing Igbo text documents. They captured issues of collocations, compounding, and word ordering that plays major roles in the language, thereby making the representation semantic-enriched. This paper is part of ongoing research on the classification of Igbo textual documents and will be used on the system. At the completion of this research, we will evaluate the performance of the three models used for the text representation, in Igbo text classification system. This can as well be adopted by any Igbo text-based work such as text mining tasks, text retrieval and natural language applications.

**Acknowledgment**
The authors wish to express gratitude the unknown reviewers of this work for their useful comments and contributions that assisted in enhancing the worth of this paper.